\def\year{2021}\relax
\documentclass[letterpaper]{article} 
\usepackage{aaai21}  
\usepackage{times}  
\usepackage{helvet} 
\usepackage{courier}  
\usepackage[hyphens]{url}  
\usepackage{graphicx} 
\usepackage{natbib}  
\usepackage{caption} 
\title{
Toward Robust Long Range Policy Transfer
}
\author{

    Wei-Cheng Tseng\textsuperscript{\rm 1}, 
    Jin-Siang Lin\textsuperscript{\rm 1}, 
    Yao-Min Feng\textsuperscript{\rm 1}, 
    Min Sun\textsuperscript{\rm 1,2,3}
}
\affiliations{

    \textsuperscript{\rm 1}National Tsing Hua University\\
    \textsuperscript{\rm 2}Appier Inc., Taiwan\\
    \textsuperscript{\rm 3}MOST Joint Research Center for AI Technology and All Vista Healthcare, Taiwan\\
    
    \{weichengtseng, linjinsiang, yaominlouis\}@gapp.nthu.edu.tw, sunmin@ee.nthu.edu.tw
}

\usepackage[export]{adjustbox}
\usepackage{wrapfig}
\usepackage[dvipsnames]{xcolor}
\usepackage[ruled,vlined]{algorithm2e}
\usepackage{lipsum}
\usepackage{ulem}
\usepackage{subcaption}
\usepackage{amsfonts}
\usepackage{amsmath}
\usepackage[switch]{lineno}
\definecolor{warning}{HTML}{000000} 
\definecolor{not_complete}{HTML}{000000}
\definecolor{advisor}{HTML}{000000}
\definecolor{ethan}{HTML}{000000}
\definecolor{special}{HTML}{000000}
\definecolor{aaai}{HTML}{000000}
\definecolor{louis}{HTML}{000000}
\definecolor{justin}{HTML}{000000}
\definecolor{eric}{HTML}{000000}
\definecolor{normal}{HTML}{000000}
\definecolor{mengli}{HTML}{000000}
\definecolor{brown}{HTML}{000000} 
\definecolor{red}{HTML}{000000}

\begin{document}
\maketitle

\begin{abstract}
    Humans can master a new task within a few trials by drawing upon skills acquired through prior experience. To mimic this capability, hierarchical models combining primitive policies learned from prior tasks have been proposed. However, these methods fall short comparing to the human's range of transferability. We propose a method, which leverages the hierarchical structure to train the combination function and adapt the set of diverse primitive polices alternatively, to efficiently produce a range of complex behaviors on challenging new tasks. We also design two regularization terms to improve the diversity and utilization rate of the primitives in the pre-training phase. We demonstrate that our method outperforms other recent policy transfer methods by combining and adapting these reusable primitives in tasks with continuous action space. The experiment results further show that our approach provides a broader transferring range. The ablation study also show the regularization terms are critical for long range policy transfer. Finally, we show that our method consistently outperforms other methods when the quality of the primitives varies.
\end{abstract}

    \section{Introduction} \label{introduction}

        Reinforcement learning (RL) has lots of success in various applications, such as game playing \cite{arxiv16_gym, nature17_alphagozero, nature15_dqn}, robotics control \cite{arxiv18_deepmind_control_suite, arxiv19_pybullet}, molecule design \cite{neurips18_gcpn}, and computer system optimization \cite{neurips19_park, iclr19_variance_reduction}. 
        Typically, researchers use RL to solve each task independently and from scratch, which makes RL confronted with low sample efficiency.
        However, compared with humans, the transferability of RL is limited. Especially, humans can learn to solve complex continuous problems (both state space and action space are continuous) efficiently by utilizing prior knowledge. In this work, we want agents to efficiently solve the complex continuous problem by exploiting prior experiences that provide structured exploration based on effective representation.
        
        To this end, we formulate transfer learning in RL as following. We train a policy with one of the RL optimization strategies on the pre-training task. Then, we intend to leverage the policy to master the transferring task. However, transfer learning in RL may face some fundamental problems. First, unlike supervised learning, the transitions and trajectories are sampled during the training phase based on the interacted policy \cite{iclr19_proml}. Since the reward distributions are different between the pre-training task and the transferring task, directly finetuning the pre-training policy on transferring tasks may make the agent perform biased structured exploration and get stuck in many low reward trajectories.
        Second, dynamics shifts between pre-training and transferring tasks may induce the pre-training policy to perform unstructured exploration \cite{iclr19_adapt_dynamics, corl19_multiagent_sim2real}. Although domain randomization \cite{iros19_domain_rand, corl19_multiagent_sim2real} in the pre-training phase may mitigate this problem, we prefer the pre-trained policies to gradually fit the transferring tasks.
        
        Some methods intend to limit dependency between pre-training policy and task-specific information \cite{iclr18_inforbot, iclr19_information_asymmetry} by using information bottleneck \cite{iclr17_vib} and variational inference \cite{iclr13_vae}. That way, that pre-training policy does not overfit to a specific task and can be transferred to other tasks. Besides, some methods achieve task transfer by embedding tasks into a latent distribution \cite{iclr18_neuralprob, iclr18_learning}.
        However, the latent distribution should be smooth and contain a diverse set of tasks to perform behavior well. Some works propose a hierarchical policy \cite{iclr17_mlsh, aaai17_option_critic, neurips19_mcp}, which contains a combination function to control how to select or combine a set of primitives. Those works acquire a new selection or combination strategy to control the primitives to master the transferring task if we attain a set of task-agnostic primitives. We find that hierarchical architecture has the potential to enable a better transferring range in continuous control problems.
        
        \textcolor{aaai}{We propose a transfer learning method in RL.} 
        Our pre-training method leverages existing hierarchical structure in the policy consisting of a combination function and a set of primitive policies. \textcolor{aaai}{We also design objectives to encourage the set of primitives to be diverse and more evenly utilized in pre-training tasks.} Notice that we do not use reference data since we expect our method to be generally applicable to all tasks. In many cases, such as flying creatures \cite{siggraphasia18_dragon}, 
        Laikago robot\footnote{http://www.unitree.cc/e/action/ShowInfo.php?classid=6\&id=1} or D’Kitty robot\footnote{https://www.trossenrobotics.com/d-kitty.aspx}, reference data is hard to obtain.
  
        \begin{figure*}[ht]
            \centering
            \begin{subfigure}{.3735\textwidth}
                \centering
                \includegraphics[width=1\linewidth]{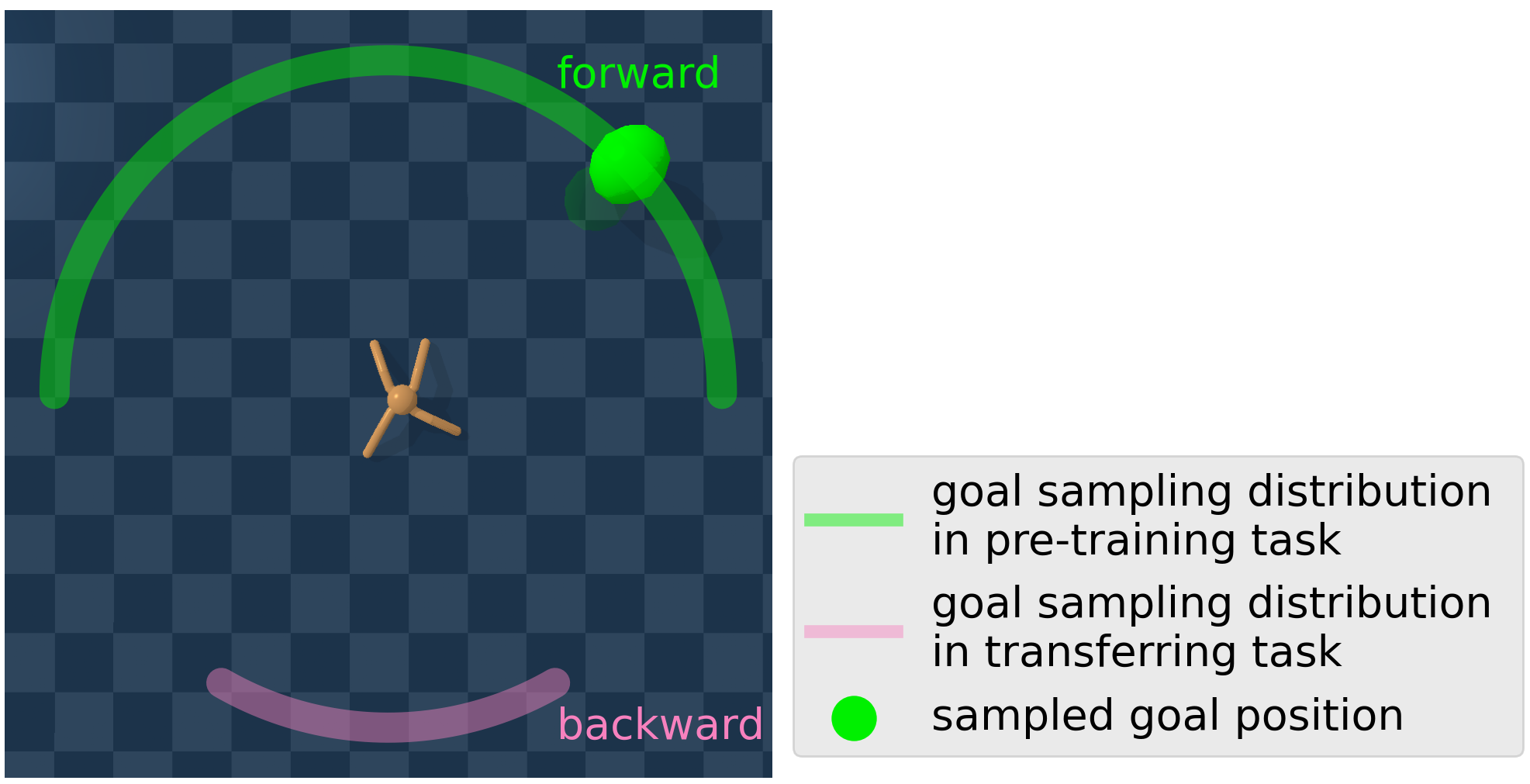}
                \caption{}
                \label{fig:motivated_example}
            \end{subfigure}
            \centering
            \begin{subfigure}{.5175\textwidth}
                \centering
                \includegraphics[width=1\linewidth]{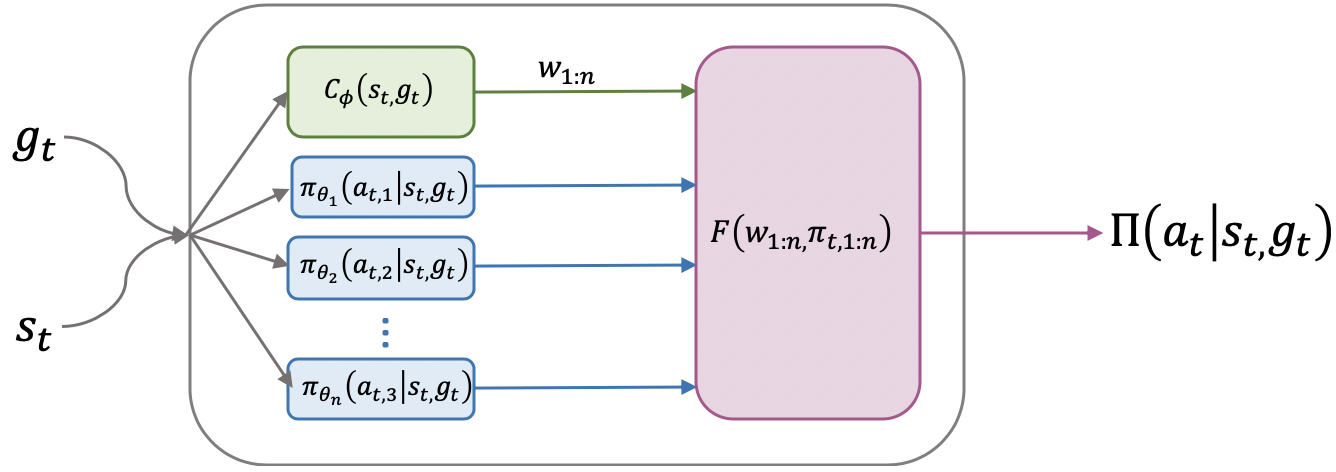}  
                \caption{}
                \label{fig:architecture}
            \end{subfigure}
            \caption{(a) Our motivating example for RL transferring. The green ball represents the target position, which is sampled from the distribution of the task. The goal direction of the pre-training task and transferring task are quite different. (b) The hierarchical policy architecture.}
            \label{fig:fig}
        \end{figure*}   
        
        During the transferring phase, we alternatively train the combination function and the primitive policies. This training procedure makes the training not only stable but also flexible in exploration. When training the combination function and freezing primitives in the transferring phase, it utilizes the benefit of the hierarchical structure that abstracts the exploration space. When training the primitives and fixing the combination function, the primitives can be adapted to the transferring task.
        In our experiments, we demonstrate that training hierarchical policy with our method significantly increases sample efficiency compared to previous work \cite{neurips19_mcp}. Moreover, our method provides a better transferring range. We also provide an ablation study to discuss the effectiveness of our regularization terms.
        Finally, we show that with different resource constraints on training the pre-training policy, our method still outperforms other methods. The source code is available to the public\footnote{https://weichengtseng.github.io/project\_website/aaai21}.

    \section{Preliminaries} \label{preliminaries}
        
        We consider a multi-task RL framework for transfer learning, consisting of a set of pre-training tasks and transferring tasks. An agent is trained from scratch on the pre-training tasks. Then, it utilizes any skills acquired during pre-training to the transferring tasks. Our objective is to obtain and leverage a set of reusable skills learned from the pre-training tasks to enable the agent to efficiently explore and be more effective at the following transferring tasks. 
        
        We denote $s$ as a state, $a$ as an action, $r$ as a reward, and $\tau$ as a trajectory consisting of actions and states. Each task is represented by a dynamics model $s_{t+1} \sim p(s_{t+1}|s_t, a_t)$ and a reward function $r_t = r(s_t, a_t, g)$, where $g$ is the task-specific goal such as the target location that an agent intends to reach and a terrain that an agent needs to pass.
        In multi-task RL, goals $\{g\}$ are sampled from a distribution $p(g)$.
        
        Given a goal $g$, a trajectory $\tau=\{s_0,a_0,s_1,...,s_T\}$ with time horizon $T$ is
        sampled from a policy $\pi(a|s,g)$. Our objective is to learn an optimal policy $\pi^{*}$
        that maximizes its expected return $J(\pi) = \mathbb{E}_{g \sim p(g),\tau \sim p_\pi (\tau|g)} [\Sigma_{t=0}^{T} \gamma^t r_t]$ over the distribution of goals $p(g)$ and
        trajectories $p_\pi (\tau|g)$, where $\gamma \in [0, 1]$ is the discount factor. 
        The probability of the trajectory $\tau$ is calculated as follow
        \begin{equation}
            p_{\pi}(\tau|g) = p(s_0)\prod_{t=0}^{T-1}p(s_{t+1}|s_t, a_t) \pi(a_t|s_t,g)
        \end{equation}
        where $p(s_0)$ is the probability of the initial state $s_0$.
        In transfer learning, despite the same state and action space, the goal distributions, reward functions, and dynamics models in pre-training and transferring tasks are subjected to be different.
        The difference between the pre-training and transferring tasks
        is referred to as the range of transfer.
        Note that a successful transfer can't be expected for totally unrelated tasks. We consider the scenario where the pre-training tasks can make the agent to learn relevant information of the following transferring tasks, but may not cover the entire set of skills which are useful at the transferring tasks.

    \section{Method} \label{method}

        We will first describe a motivating example in Sec. \ref{motivated_example}. Then we introduce our method in Sec. \ref{our_method}. Finally, we show how to apply our method to the existing hierarchical policy in Sec. \ref{apply_hat}.
        \subsection{A Motivating Example} \label{motivated_example}

            Let's consider a model-free RL scenario. We have a pre-training task that an ant needs to achieve a goal position to get some reward, and the goal position is sampled from a half-circle in front of the ant.
            However, we change the sample distribution of goal positions in the transferring task to a small arc in the back of the ant that does not overlap with the goal positions of the pre-training task (see Fig. \ref{fig:motivated_example}). Intuitively speaking, this is a challenging transferring scenario.

            There are two straightforward methods to tackle this problem. One is directly finetuning the pre-training policy on the transferring task. However, the pre-trained policy is affected by the goal distribution in the pre-training phase to move forward, and this conflicts with the goal distribution in transferring task to move backward. Therefore, directly finetuning may corrupt the information learned from the pre-training task, which is known as the catastrophic forgetting. The other is to train a new policy from scratch on the transferring task. Training from scratch can saturate to a good point, but it may need lots of trails. Our proposed method aims to address the drawbacks of these two basic methods.
            
        \subsection{Our Method} \label{our_method}
            \textcolor{aaai}{
            We describe our method in this section. Our policy architecture contains: 1) A set of primitive policies $\pi_{\theta_1}(a|s, g), \pi_{\theta_2}(a|s, g), ..., \pi_{\theta_n}(a|s, g)$ with parameters $\theta_{1:n}$, and each primitive is an independent policy that outputs action distribution conditioned on $s$ and $g$. 2) A combination function $C_{\phi}(s, g)$ with parameter $\phi$ outputs weight $w_{i:n}$, where $w_i$ specifies the importance of primitive $\pi_{\phi_i}$. $F(\pi_{1:n}, w_{1:n})$ specifies how to combine primitives with specified weight $w_{1:n}$. Typically, the larger the weight $w_i$, the more contributions from primitive $\pi_{\theta_i}$.
            }
            
            \textcolor{aaai}{
            During the pre-training phase, the hierarchical policy is end-to-end trained with an off-the-shelf RL optimization method. Our goal is to learn a set of primitives such that the combination function can compose them to form a complete behavior. However, not all primitives learned in pre-training are equally good for long range transfer. We identify two issues to be addressed. First, diversity of primitives is essential to improve the transferability. Without encouraging diversity, some of the primitives may learn similar behavior which better fits the pre-training tasks but affects the transferability. Second, when more primitives are introduced to compose more complex behaviors and improve the transferability, the utilization rate of each primitive varies more. In some extreme cases, some primitives may seldom or never used in the pre-training phase, so these primitives are not updated by RL optimization. Hence, it is important to encourage the utilization rates of primitives to be more evenly distributed.}
            
            \begin{algorithm}
                \SetCustomAlgoRuledWidth{0.535\textwidth} 
                \caption{Full Algorithm}
                    // pre-training phase \;
                    initialize $\theta_{1:n}$ and $\phi$ \;
                    let $J_{RL}(\theta_{1:k}, \phi)$ be the objective function of some specific RL optimization \;
                    \While{not converge}{
                    train combination function $\phi$ and primitives $\theta_{1:n}$ with
                    $J_{pre}(\theta_{1:k}, \phi) = J_{RL}(\theta_{1:k}, \phi) + \alpha \cdot J(\pi_{1:k}) + \beta \cdot J(w_{1:k})$
                    }
                    \texttt{\\}
                    \texttt{\\}
                    // transferring phase \;
                    reinitialize $\phi$ \;
                    \While{not converge}{
                        Disable the gradient of primitives \;
                        Enable the gradient of combination function \;
                        \For{$i = 1: p$}{
                            train combination function $\phi$ with \textcolor{aaai}{$J_{transfer}(\theta_{1:k}, \phi) =J_{RL}(\theta_{1:k}, \phi)$}
                        }
                        Disable the gradient of combination function \;
                        Enable the gradient of primitives \;
                        \For{$i = 1: p$}{
                            train primitives $\theta_{1:n}$ with \textcolor{aaai}{$J_{transfer}(\theta_{1:k}, \phi) =J_{RL}(\theta_{1:k}, \phi)$}
                        }
                    }
                    \label{algorithm:hat}
            \end{algorithm}
            
            \textcolor{aaai}{
            To mitigate these two issues, we propose two regularization terms. We introduce the first regularization term that \textcolor{louis}{separates the distributions of the} primitives from each other's so that the primitives become diverse. The intuitive idea to measure the difference between two probability distributions is calculating KL divergence. Therefore, we calculate the average of arbitrary pair of primitives as our regularization term, and we call it Diversity Regularization (DR)
            }
            \begin{equation}
                \begin{split}
                    J(\pi_{\theta_{1:k}}) & = \frac{1}{k(k-1)} \sum_{i \ne j}{D_{KL}(\pi_{\theta_i} | \pi_{\theta_j})} \\ 
                    & = \frac{2}{k(k-1)} \sum_{i \ne j, i<j}{D_{JS}(\pi_{\theta_i} | \pi_{\theta_j})} 
                \end{split}
            \end{equation}
            \textcolor{aaai}{
            Then, we propose another regularization term to encourage utilization rates of primitives \textcolor{louis}{to be} more evenly used in the pre-training task. We try to model the weight of primitives as categorical distribution and use the entropy of distribution as our regularization term, and we call it Utility Regularization (UR).
            }
            \begin{equation}
                \begin{split}
                    J(w_{1:k}) = -\sum_{i}{(\frac{e^{w_i}}{\sum_{j}{e^{w_j}}})\log\frac{e^{w_i}}{\sum_{j}{e^{w_j}}} }
                \end{split}
            \end{equation}
            
            \textcolor{aaai}{
            Therefore, the overall objective in the pre-training phase is shown as below.
            }
            \begin{equation}
                J_{pre}(\theta_{1:k}, \phi) = J_{RL}(\theta_{1:k}, \phi) + \alpha \cdot J(\pi_{1:k}) + \beta \cdot J(w_{1:k})
            \end{equation}
            \textcolor{aaai}{
            where $\alpha$ and $\beta$ are the hyperparameters and $J_{RL}(\theta_{1:k}, \phi)$ can be any reinforcement learning optimization method.
            }
            
            During the transferring phase, we reinitialize the weights of the combination function and leverage the primitive directly from pre-trained in the pre-training task. We first update the combination function and freeze the primitives. This views the combination function as the policy that learns to combine the primitives to master the transferring task. As illustrated in the motivating example, the range of transfer between the pre-training and transferring tasks is likely to limit the performance of only training the combination function. Therefore, after $p$ iterations, we switch to finetune the primitives with the combination function fixed. This method makes the primitives become more applicable to the transferring task. After $p$ iterations, we freeze primitives and train combination function again. This is to prevent the skills in the primitives to be severely forgotten during finetuning. The strategy is repeated several times until the hierarchical policy converges (see algorithm 1).

        \subsection{Applying Our Method to the Policy Architecture} \label{apply_hat}
            We leverage the multiplicative combination rule \cite{neurips19_mcp}
            \begin{equation}
                F(\pi_{\theta_{1:n}}, w_{1:n}) = \frac{1}{Z(s,g)} \prod_{i=1}^{k} \pi_{\theta_i}(a|s)^{w_i}
            \end{equation}
            where $\pi_{\theta_{1:n}}(a|s)$ is the primitive policies and $w_{1:n}$ is generated from combination function $C_{\phi}(s, g)$. $Z(s, g)$ is the partition function that ensures the composite distribution is normalized.
            $F(\pi_{\theta_{1:n}}, w_{1:n})$ multiplies the primitive policies along with their corresponding weights. The weights determine the importance of each primitive policies to compose the action distribution at a time step,
            with a larger weight representing a larger influence. 
            Note that to make the primitives task-agnostic (i.e., more transferable), we restrict the primitives $\pi_{\theta_{1:n}}(a|s)$ to only get $s$.
            During the pre-training phase, the combination function and primitive policies are trained in an end-to-end manner. During the transferring phase, we train the combination function and primitive policies alternatively as described in the previous section.

    \section{Related Works} \label{related_works}

        Learning meaningful and reusable representations that can be transferred across multiple tasks is a popular research direction in machine learning \cite{neurips06_multitask}. One of the straightforward transfer learning methods is finetuning. That is, a network is first trained on a source domain. Then, the learned representation or features are reused in another domain by finetuning via backpropagation \cite{science06_reduce, aaai18_measure, arxiv16_overcome}. However, backpropagation may destroy previously learned representations or features before the network leverages them in the target domain.
        
        \subsection{Information-Based Methods}
            Some methods try to learn a policy that doesn't highly depend on task-specific information during pre-training and only acquire task-specific information when the policy needs to make a critical decision to solve the task \cite{iclr18_inforbot, iclr19_information_asymmetry, iclr20_information_constrain}. These methods may leverage some concepts like information bottleneck \cite{iclr17_information_bottleneck} and variational inference \cite{iclr13_vae} to restrict policy and task-specific information dependency. Since the policy only gets limited task-specific information, the policy can potentially be transferred to other tasks.
            
        \subsection{Hierarchical Methods}
            Some methods intend to use a hierarchical policy that typically contains a master policy or combination function and a set of low-level policies \cite{aaai17_option_critic}. Then, by training the hierarchical policy in the pre-training task and acquire a set of primitives, we can reschedule or recompose these to master transferring tasks. The combination function in some method only activates one sub-policy at a time \cite{siggraph17_schedule_fragments, siggraph18_deepmimic, iclr17_mlsh} while the other combination function allows multiple primitive to be executed at the same time \cite{neurips19_mcp, iclr20_composing}.
            \textcolor{aaai}{Another kind of methods leverages some ideas from hierarchical reinforcement learning (HRL) \cite{neurips18_hiro,iclr18_multilayer_hier}. The policy architecture typically contain a high-level policy and a low-level policy. The high-level policy outputs a goal state or condition to the low-level policy, and the low-level policy intends to achieve the given goal state. Some works \cite{neurips19_haar,iclr20_subpolicy_adapt} propose to finetune policy with HRL framework to perform transfer learning in RL.} 
            
        \subsection{Latent Space Methods}
            Some methods specify actions through the latent representation which is further transformed to the actions of the underlying system \cite{iclr18_neuralprob, robotics08_latent, icml19_action_representation}. Therefore, a common approach is getting the latent space by pre-training the policy on the pre-training task, and further transfer the policy to the downstream task \cite{icml18_latent_space_policy}. To encourage the latent space is diverse enough, one solution is to leverage reference data \cite{sfu_mocap, cmu_mocap} to pre-train the latent space \cite{iclr18_neuralprob}. Other diversity-driven pre-training methods have been proposed to help the latent space to form semantically various behaviors \cite{iclr18_learning}.
      
        \begin{figure}[t!]
            \centering
            \includegraphics[width=.43\textwidth]{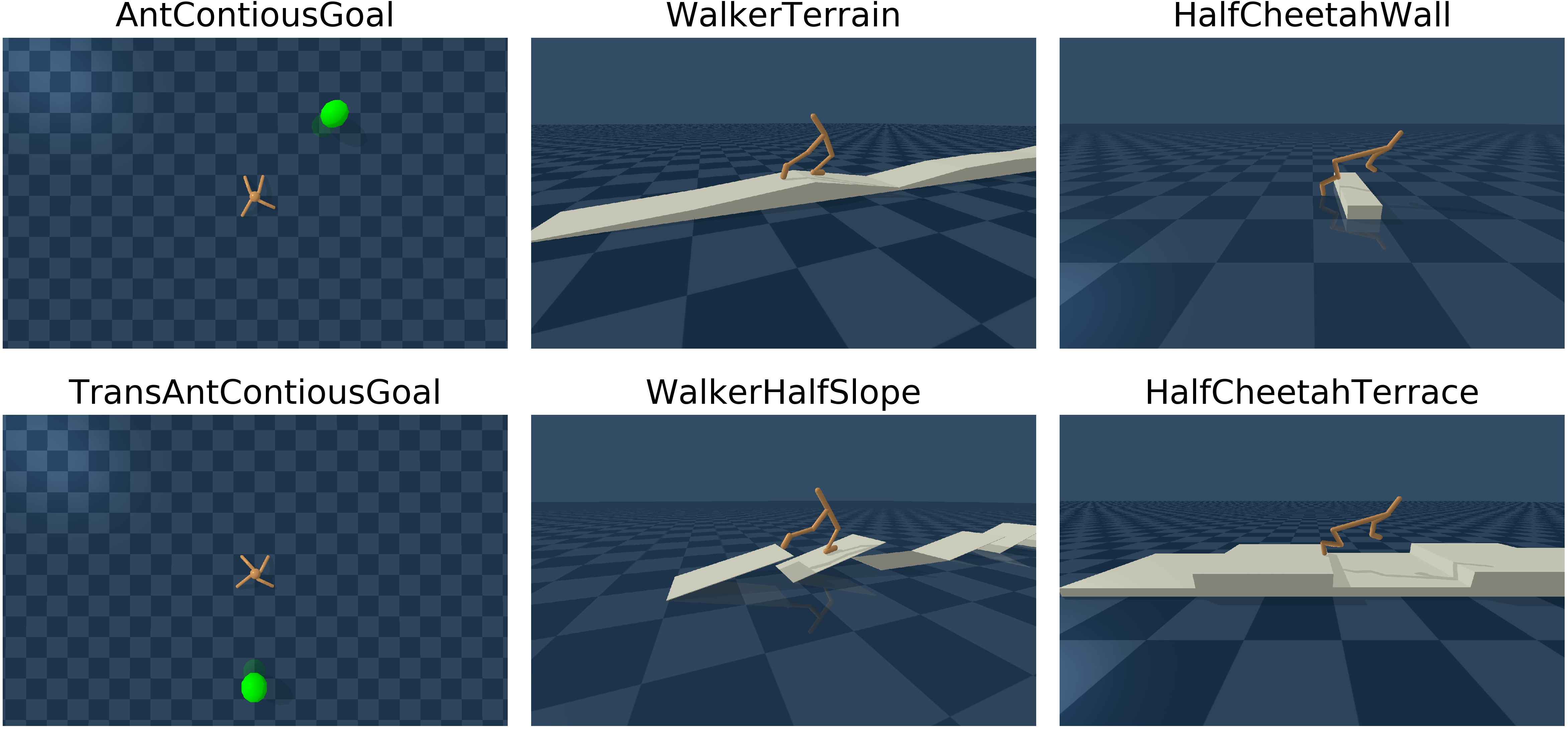}
            \caption{Environments used to evaluate our method. The first row is the pre-training tasks, and the second row is the transferring tasks. All the baseline and our method are evaluated with the three agents: ant, 2dwalker, halfcheetah.}
            \label{fig:environments}
        \end{figure}    
            
    \section{Experiments} \label{experiments}
        \begin{figure*}[t]
            \centering
            \includegraphics[width=0.9\textwidth]{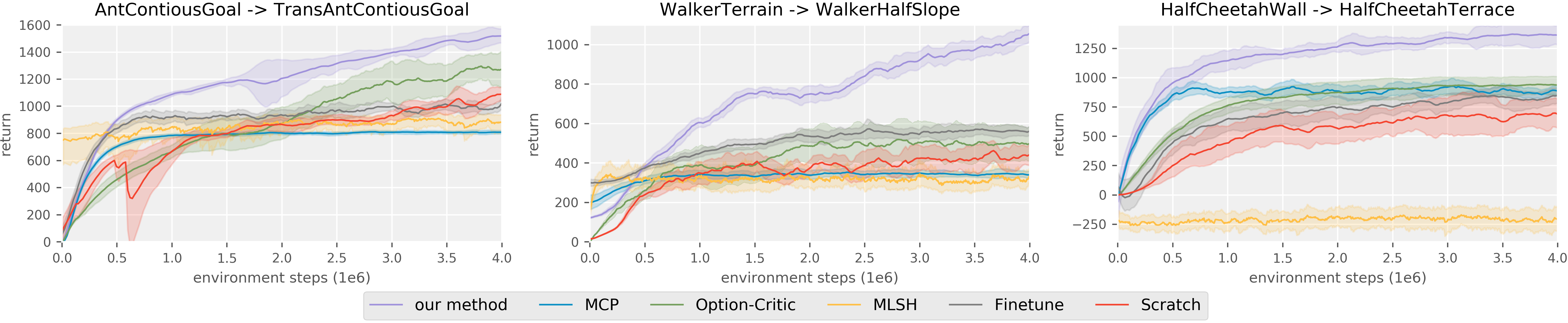}
            \caption{Performance of different transferring methods. For better visualization, we use exponential moving average to smooth the learning curve, and each learning curve is grouped with ten random seeds. \textcolor{ethan}{The transparent part represents the maximum and minimum of the learning curves.} From these figures, we show that our method achieves better performance than other methods.
            }
            \label{fig:performance}
        \end{figure*}

        In this section, we introduce the evaluation tasks in Sec. \ref{task} and list the baselines that we intend to compare in Sec. \ref{baselines}. The results of the methods evaluated in our environments are discussed in Sec. \ref{comparisions}. Aside from option-critic \cite{aaai17_option_critic}, all the experiments are trained with PPO \cite{arxiv17_ppo} and Generalized Advantage Estimation (GAE) \cite{iclr16_gae}. 
        The detailed hyperparameter settings are shown in supplementary Sec. 4.1. We further show that our method has a broader transferring range compared with other baselines in Sec. \ref{transfer_range}. 
        \textcolor{aaai}{Then, we demonstrate the effectiveness of the regularization terms in Sec. \ref{ablation}.} 
        Finally, we demonstrate that our method can perform well even if the primitive policies are not good enough in Sec. \ref{effectiveness_of_pretrain_policy}. \textcolor{ethan}{Some addition proprieties of our method are further discussed in supplementary Sec. 2.}

        \subsection{Tasks} \label{task}
            We consider three agents (see Fig. \ref{fig:environments}): quadruped (ant) with 12 DoF and 8 actuators; 2dwalker with 6 DoF and 6 actuators; 
            halfcheetah with 6 DoF and 6 actuators. 
            All the tasks are built with PyBullet \cite{arxiv19_pybullet}. The implementation detail of these tasks will be described in supplementary Sec. 4.2. and we discuss the tasks in the following sections.

            \subsubsection{Pre-training tasks} \label{pretrain_task}
            \begin{itemize}
                \item \textbf{AntContinuousGoal}: An ant needs to move to the target position which is the task-specific information $g$, and the target direction is sampled from [-0.5$\pi$, 0.5$\pi$] with radius 5 as we showed in Fig. \ref{fig:motivated_example}.  Once the ant reaches the goal position, the goal position will be re-sampled in the same way.
                
                \item \textbf{WalkerTerrain}: A 2d walker needs to move forward on terrain, and the slope of the terrain is sampled from a specific range. The task-specific information $g$ is the terrain in front of the agent. Therefore, the 2d walker needs to learn how to walk smoothly on planes with different slopes.
                
                \item \textbf{HalfCheetahWall}: A halfcheetah needs to move forward, and there is a wall every 3 to 5 meters. The height of the wall is sampled from a specific range. The task-specific information $g$ is the terrain in front of the agent. Therefore, the halfcheetah needs to climb or jump over the wall.            
            \end{itemize}
            \subsubsection{Transferring tasks} \label{transfer_task}
            \begin{itemize}
                \item \textbf{TransAntContinuousGoal}: An ant needs to move to the target position, and the target position is sampled from [5$\pi$/6, 7$\pi$/6] with radius 5, as we showed in Fig. \ref{fig:motivated_example}. Note that the target direction does not overlap with that of the pre-training task. Once the ant reaches the goal position, the goal position will be re-sampled in the same way. Therefore, in this case, the difference between pre-training task and transferring task is goal distribution $p(g)$.

                \item \textbf{WalkerHalfSlope}: A 2d walker needs to move forward in terrain, but there are cliffs between each plane. Therefore, the 2d walker should be robust to these cliffs. Therefore, in this case, the differences are goal distribution $p(g)$ and dynamics $p(s_{t+1}|s_t, a_t)$.
        
                \item \textbf{HalfCheetahTerrace}: 
                A halfcheetah needs to move forward on a terrace that is formed with lots of horizontal platforms with target speed, so it needs to jump or climb up to the higher platform and does not fall to the lower platform. The difference in height between two platforms is sampled from a specific range. Therefore, in this case, the differences are goal distribution $p(g)$ and dynamics $p(s_{t+1}|s_t, a_t)$.
        \end{itemize}
        
        \subsection{Baselines} \label{baselines}

            \begin{figure*}[t]
                \centering
                \includegraphics[width=0.9\textwidth]{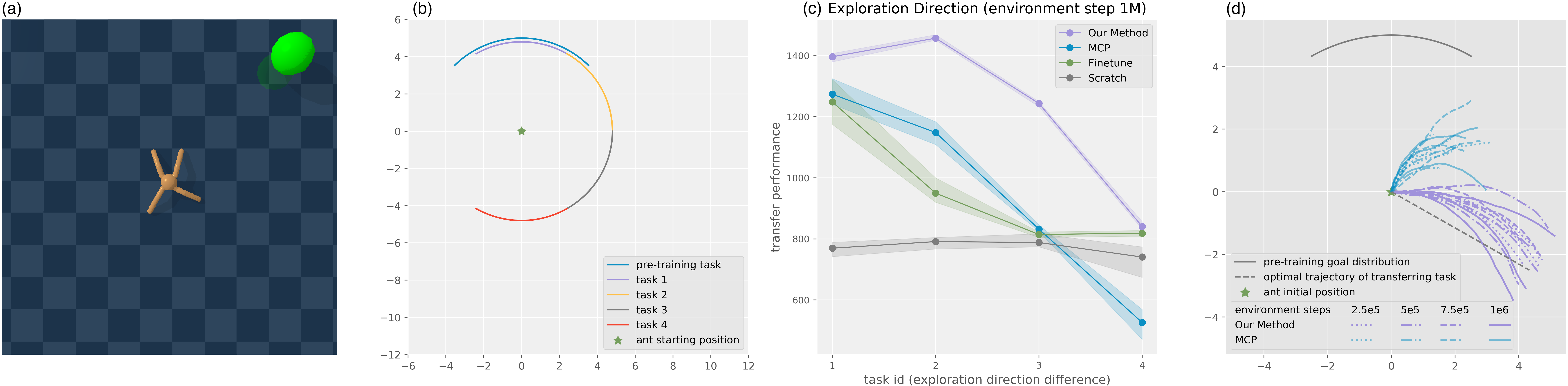}
                \caption{Performance of different transferring methods. From these figures, we show that our method achieves better performance than other methods. (a): the experiment is based on the ant agent. (b): the goal position sampling range for each task. From task 1 to task 4, the task difference between transferring task and pre-training task becomes larger. (c): the result of each method on different transferring tasks. (d): the trajectories of our method and MCP under different numbers of training iterations. It is clear that our method was not affected by the pre-training task compared with MCP.}
                \label{exploration_direction}
            \end{figure*}
        
            We define the baselines that will be discussed in the following sections, and the implementation detail will be described in supplementation material Sec. 4.1.
       
            \textbf{Scratch}: We directly train a policy with PPO\cite{arxiv17_ppo} on the transferring task, and it is one of the most straightforward methods to tackle a task.
            \textbf{Finetune}: We first train a policy in the pre-training task. Then, we directly finetune the policy in the transferring task.
            It is another the most straightforward method to tackle a task.
            \textbf{MCP} \cite{neurips19_mcp}: A multiplicative model that enables the agent to activate multiple primitives simultaneously is trained. Each primitive specializes in different behaviors that can be combined to span a continuous spectrum of skills on the pre-training task. During the transferring phase, the primitives are fixed, and only the combination function is updated.
            \textbf{Option-Critic} \cite{aaai17_option_critic}: Both intra-option policies and termination conditions of options are learned, followed by the policy over options, and without specifying any additional rewards or subgoals. The options are assigned to perform sequentially. One option works for several timesteps until being stopped by the termination function. The entire network is pre-trained on the pre-training task. Then, it is directly finetuned to the transferring task.
            \textbf{MLSH} \cite{iclr17_mlsh}: \textcolor{justin}{A hierarchical policy where the master policy switches between a set of sub-policies is learned. The master policy chooses a sub-policy every N timesteps, and the selected sub-policy is then executed by the agent for N timesteps to interact with the environment to constitute a high-level action. Both master policy and sub-policies are optimized on the pre-training task. During the transferring phase, we freeze the sub-policies and only train the master policy, which makes the master policy learn how to utilize the fixed primitives.}

            Aside from the baselines mentioned above, some additional baselines are discussed in supplementary Sec. 1.

        \subsection{Comparisons with Baseline} \label{comparisions}
            We run our method and other baselines in the three continuous problems described in the previous section. The network architecture, hyperparameter setting, and other implementation detail will be described in supplementary Sec. 4.1. For Fig. \ref{fig:performance}, we find that our method outperforms other methods in the transferring phase. MCP tends to perform well at the beginning of training, and this phenomenon may be caused by hierarchical abstraction. However, freezing primitive policies may restrict the transferring range of MCP, which 
            induces that MCP cannot converge to a better result. Since our method acquires a set of distinct and even utilization rate primitives and allows the primitives to adapt to the transferring task, it achieves significantly better reward the fastest among all the baseline algorithms. \textcolor{louis}{With the knowledge learned from the pretraining tasks, finetuning performs better than training from scratch. This is because training from scratch is required to learn everything, including task distribution and dynamics. However, our method significantly outperforms finetuning by striking a better trade-off between combining the primitives to efficiently exploring the new task and gradually adapting the primitive to the new task.}
            \textcolor{justin}{MLSH and option-critic do not perform well in our transfer task since \textcolor{louis}{they} chooses the primitives serially, which makes the primitives not decomposed well. When the task distribution and dynamics change in the transferring phase, They are no longer able to provide suitable behavior.}

\begin{figure}
                \centering
                \includegraphics[width=0.4\textwidth]{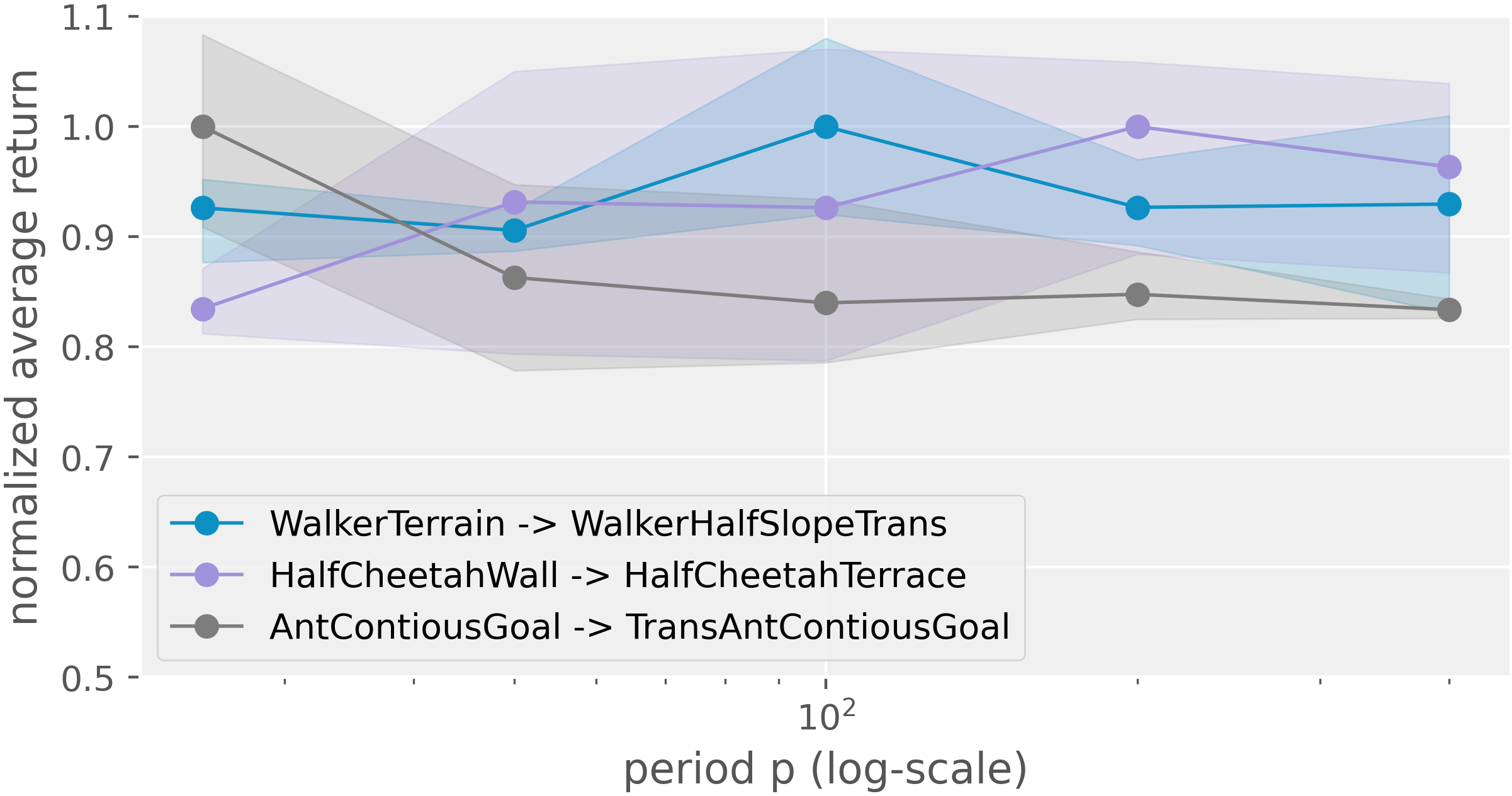}
                \caption{Normalized performance with different period $p$. 
                The performance is normalized by the best average return achieved in each task so that we can plot the result of all the tasks in a single plot.
                We find that the performance is not affected much by the period $p$. 
                }
                \label{fig:different_period}
            \end{figure}

            One concern about our method is that it introduces a hyperparameter $p$, which controls the period of 
            the number of the gradient steps before switching between primitives and combination function.
            To this end, we show the performance of our method with different period $p$ in Fig. \ref{fig:different_period}. We find that our method is not sensitive to period $p$, so there is not much effort to tune this hyperparameter.

        \subsection{Transferring Range} \label{transfer_range}
        
            The transferring range is critical in transfer learning. To demonstrate the transferring range of each method, we redesign the goal position of AntContinuousGoalEnv (see Fig. \ref{exploration_direction} (a, b)). The goal position pre-training task is sampled from an arc where the center angle is [$-\pi / 4$, $\pi / 4$] and radius 5 meters. 
            As for transferring tasks, we design four transferring tasks, and the goal position of these four transferring tasks are sampled from [$-\pi / 6$, $\pi / 6$], [$\pi / 6$, $3\pi / 6$], [$3\pi / 6$, $5\pi / 6$] and [$5\pi / 6$, $7\pi / 6$], respectively. In other words, the four transferring tasks are ordered by the scale of the task difference between the pre-training task and the transferring task.
            
            \begin{figure*}
                \centering
                \includegraphics[width=0.92\textwidth]{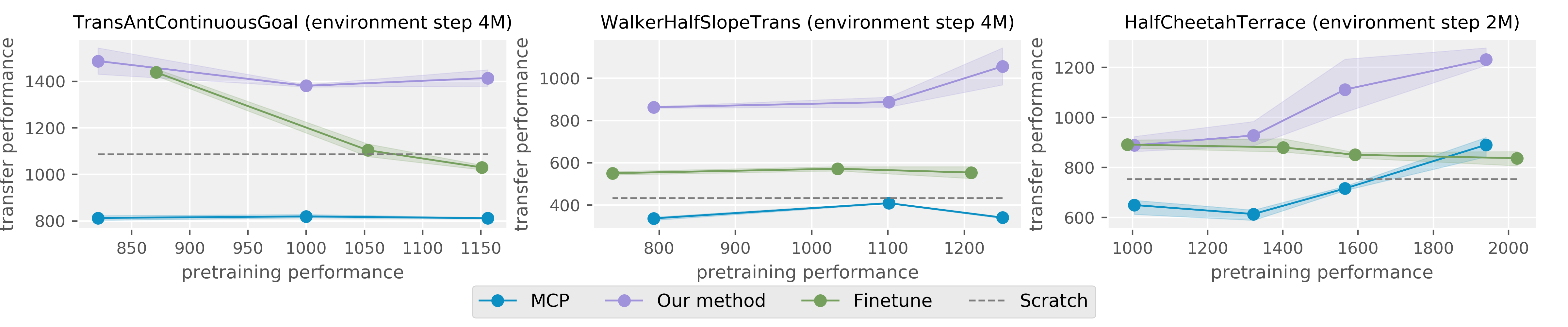}
                \caption{We evaluate the effectiveness of the pre-trained policy by using different checkpoints with different performance. We find that our method outperforms other methods 
                given different pre-training performances.}
                \label{fig:pretrain_and_transfer}
            \end{figure*}
            
            In Fig. \textcolor{louis}{\ref{exploration_direction} (c)}, 
            We find that our method has a larger transferring range compared with other methods. Note that we report the performance at 1 million environment steps. All the other transferring methods get worse as exploration direction becomes different. 
            As the exploration direction difference increases, MCP may perform worse than finetuning.
            It may be caused by fixing the primitive policies, which may limit the ability to adapt to the transferring task. Besides, if we plot the trajectory of the ant in the transferring task 3 (see Fig. \ref{exploration_direction} (d)) with a fixed goal position, we find MCP is biased by the pre-training task. Though the ant mitigates this bias by more training iteration, it tends to move forward and turn to the goal direction. In contrast, since our method allows the primitive policy to adapt to transferring task, the ant tends to act in goal direction directly. The experiment result implies that our method has a better transferring range than other methods.
            
        \subsection{Ablation Study} \label{ablation}
            \begin{figure}
                    \centering
                    \includegraphics[width=0.43\textwidth]{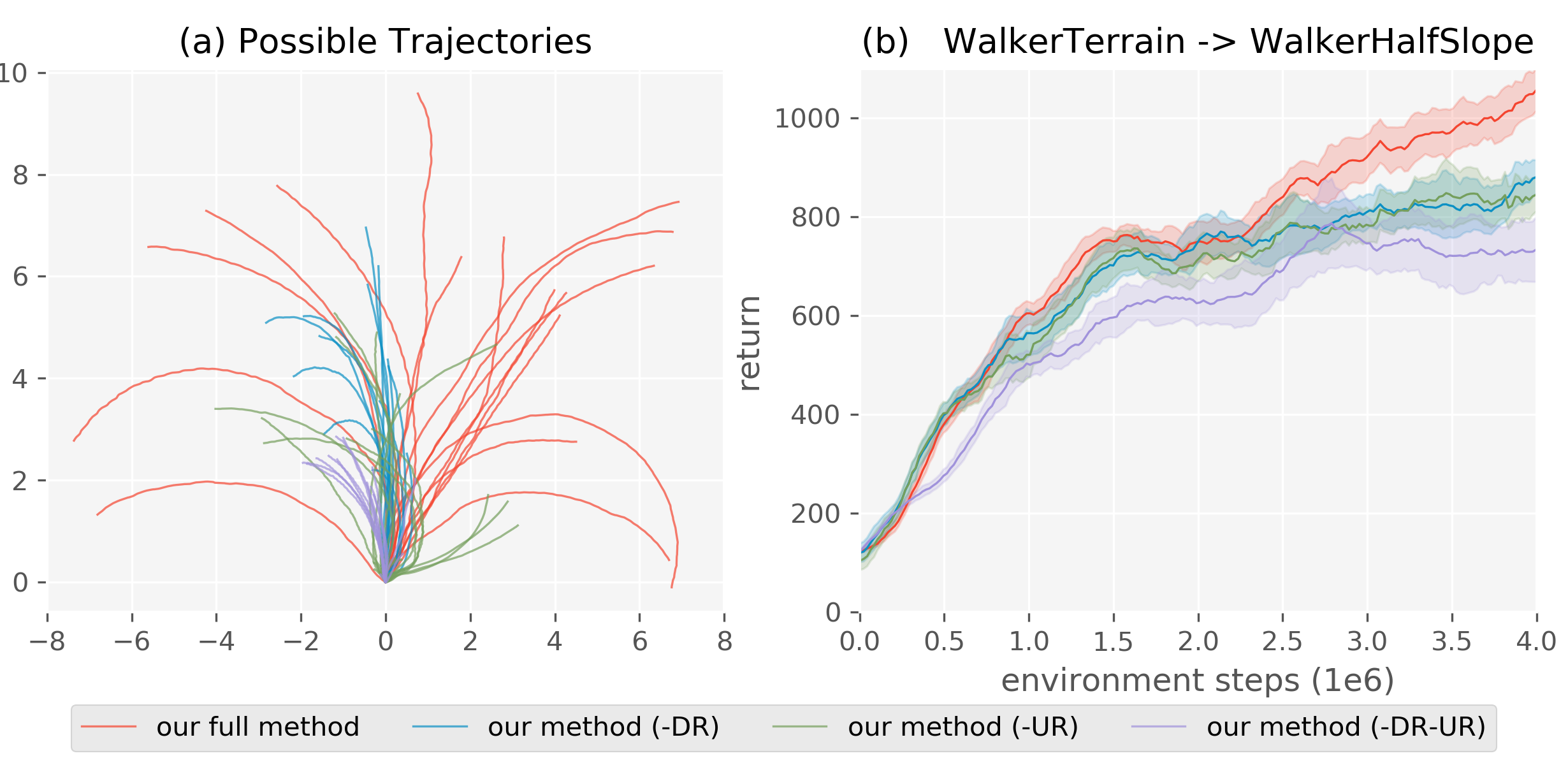}
                    \caption{Ablation study on our method. (a) Trajectories that generate with ramdom sample $w_{1:k}$. (b) We evaluate the our method with and without the regularization terms in the walker scenario.}
                    \label{fig:ablation}
            \end{figure}
            
            \textcolor{aaai}{
            To evaluate whether the regularization terms (DR and UR) are effective, we conduct the ablation study on the ant and walker scenario. For the ant scenario, we pre-train the primitives with and without the regularization terms. Then, we plot the trajectories with random sampled combination weights $w_{1:k}$ in Fig. \ref{fig:ablation}. We can find that primitives trained with both DR and UR help the agent move in a different direction with longer distance. 
            To be more specific, the trajectories with only DR tend to be short but diverse in the moving direction. It is because DR makes the primitive difference between each other. However, since DR does not enhance the utilization rate of primitives, some primitives become redundant and affect the entire policy's performance (i.e., shorter trajectories).
            On the other hand, the trajectories with only UR tend to be long but lack of diversity in the moving direction. 
            \textcolor{special}{
            Note that the long trajectories are generated even with random weights in the combination function. This is because, with UR, RL optimization more evenly updates all the primitives during pre-training.
            With both DR and UR, long and diverse trajectories enables structure exploration which helps the combination function to be efficiently retrained during the transferring phase. 
            }
            }
            
            \textcolor{aaai}{We further conduct pre-training and transferring experiments in walker scenario.
            From Fig. \ref{fig:ablation} (b), we can find that both diversity regularization and utility regularization can the policy achieve better performance with better sample efficiency.
            Our method with all regularization terms outperform other versions. \textcolor{special}{More ablation study are provide in supplementary Sec. 1.}
            }
        \subsection{Quality of the Primitives} \label{effectiveness_of_pretrain_policy}

            We further study how the quality of the primitives on the pre-training task will affect policy transfer performance. A stable policy transfer method should still perform reasonably well even when the quality of the primitives varies.
            To be more specific, we train the primitives with different training iteration in pre-training. From Fig. \ref{fig:pretrain_and_transfer}, we observe that our method achieves better performance in the transferring phase compared to other baselines. The performance of MCP is significantly affected by the quality of the primitives as the primitives are fixed. We also observed that, in the second and third tasks, the better the primitives, the better the our method performance in the transferring task. However, in the first task where the goal distributions are the opposite between the pre-training and transferring tasks, the better the primitives do not imply the better the our method performance in the transferring task. Although we found that we prefer the different quality of the primitives depending on the range of transfer, our method is the most stable method outperforming all baseline methods consistently.
            
    \section{Conclusion} \label{conclusion}
            We propose a method that leverages hierarchical structures by training over different function combinations with two novel regularization terms (DR and UR) and adapting primitive policies alternatively. The experiments show that our approach outperforms previous policy transfer methods under the long range transferring scenario for continuous action spaces. We also empirically show that our method provides a larger transferring range as well as an effective adaptation by varying the scale of the task difference between the pre-training task and transferring task. The ablation study also provides evidence of the effectiveness of our regularization terms. Finally, our method achieves more stable performance than other methods do when the quality of the primitive varies.
            
    \section{Acknowledgments}
    This project is funded by Ministry of Science and Technology of Taiwan (MOST 109-2634-F-007-016, MOST Joint Research Center for AI Technology and All Vista Healthcare).

\bibliography{reference}
\end{document}